\documentclass[sigconf, screen]{acmart}
\usepackage{subcaption}
\usepackage{booktabs}

\AtBeginDocument{%
  \providecommand\BibTeX{{%
    \normalfont B\kern-0.5em{\scshape i\kern-0.25em b}\kern-0.8em\TeX}}}

\copyrightyear{2021}
\acmYear{2021}
\setcopyright{acmlicensed}\acmConference[FDG'21]{The 16th International Conference on the Foundations of Digital Games (FDG) 2021}{August 3--6, 2021}{Montreal, QC, Canada}
\acmBooktitle{The 16th International Conference on the Foundations of Digital Games (FDG) 2021 (FDG'21), August 3--6, 2021, Montreal, QC, Canada}
\acmPrice{15.00}
\acmDOI{10.1145/3472538.3472587}
\acmISBN{978-1-4503-8422-3/21/08}

\begin{document}

\title{Adversarial Random Forest Classifier for Automated Game Design}


\author{Thomas Maurer}
\affiliation{%
  \institution{University of Alberta}
  \streetaddress{Anon Address}
  \city{Edmonton}
  \country{Canada}}
\email{tmaurer@ualberta.ca}

\author{Matthew Guzdial}
\affiliation{%
  \institution{University of Alberta}
  \streetaddress{Anon Address}
  \city{Edmonton}
  \country{Canada}}
\email{guzdial@ualberta.ca}

\renewcommand{\shortauthors}{Anon et al.}

\begin{abstract}
Autonomous game design, generating games algorithmically, has been a longtime goal within the technical games research field.
However, existing autonomous game design systems have relied in large part on human-authoring for game design knowledge, such as fitness functions in search-based methods. 
In this paper, we describe an experiment to attempt to learn a human-like fitness function for autonomous game design in an adversarial manner. 
While our experimental work did not meet our expectations, we present an analysis of our system and results that we hope will be informative to future autonomous game design research.
\end{abstract}




\maketitle


\section{Introduction}

Traditionally, game development has been a lengthy and expensive process where each piece of content is hand-authored by game designers and developers. 
Some prior research has attempted to address this through the creation of game content algorithmically. 
These algorithms have been grouped together under the term procedural content generation (PCG).
Most PCG algorithms are hand-coded, using approaches like genetic algorithms \cite{Moghadam2017} or constraint-based solvers \cite{summerville2018gemini}. 
More recently, attempts to combine machine learning (ML) with PCG have gained popularity \cite{Summerville2016,risi2020increasing}.
Procedural Content Generation via Machine Learning (PCGML) removes more overhead from the developers, requires less human authoring than traditional PCG, and can create high quality content \cite{summerville2018procedural}.
However, there has been relatively little work on using PCGML for generating entire games.
The creation of a PCGML system that can generate entire games could allow for easier adoption of this technology by practitioners and hobbyists.\par

Currently the space of PCGML game generation is under-explored.
Researchers have proposed PCGML game generators \cite{osborn2017automated,9231927}, but only a few systems have been built that can generate game mechanics, and these generally focus on specific mechanics or genres \cite{summerville2020extracting,9356343}. 
This lack of implementation is due, in part, to the inherent complexity of games and the rules that define them.
Machine learning approaches struggle with problems with high variance, and low training data such as autonomous game design, which makes them harder to implement.
The task of autonomous game generation requires both autonomous level design, and autonomous rule design at minimum. 
Machine learning has been used for PCG but generally focuses on level generation\cite{summerville2018procedural}. 
The problem of generating levels through PCGML is often easier as the representation of the data is simpler, and there is a much larger dataset. 
The application of a single PCGML system to generating both the rules of a game, and the levels for that game, is relatively unexplored \cite{9356343}.  

The most common PCG approach to automated game design has been search-based PCG \cite{browne2010evolutionary, cook2012aesthetic, cook2016angelina}.
One limiting aspect of search-based PCG is the requirement of a hand-authored fitness function.
If we could learn this fitness function from existing games, an automated measure of human-like games, we could remove this limitation and lead to more general automated game design systems.
In this work, we attempted to address this problem by replacing the hand-authored fitness function in a greedy search system, with a random forest classifier. 
Specifically, we trained a random forest classifier to differentiate between existing human-authored games and randomly generated games. 
Inspired by adversarial training approaches, we iteratively retrained our random forest classifier on the games generated by using the classifier as a fitness function.
While the games that were created by this process did not meet our expectations of human quality, they still produced interesting results.
The classifier did seem to learn some measure of structure as the games produced had patterns in level and rule design. 
While this system did not provide the results we had hoped for, there were noticeable differences in pattern consistency over the iterations of our system design that could be leveraged in future autonomous game design work.

In this paper, we present a system for autonomous game generation using PCGML, by guiding search with a random forest classifier. 
We discuss the process of iteratively training this system on its output, and present our contributions to the field of PCGML which include:
\begin{itemize}
    \item A novel representation method for VGDL games
    \item A novel method for approximating human-like fitness functions for VGDL games
    \item Experiments on this method with unexpected results
    \item An analysis of our experiments, and their implications for future work
\end{itemize}

\section{Related Work}



In this section we discuss prior work on PCG and autonomous game design.
We also detail previous PCGML approaches, particularly for autonomous game design.

\subsection{Procedural Content Generation}
Procedural content generation has been used for a wide variety of tasks, especially in game design \cite{shaker2016procedural}.
However, the majority of PCG methods rely on either constructive or search-based systems \cite{togelius2011search}.
For example, when the initial framework was presented for the General Video Game Artificial Intelligence (GVGAI) Level Generation competition, it included 2 example generators: a constructive system, and a search-based system \cite{khalifa2016general}.
We employ the GVGAI framework in our research, but our search-based system uses a machine learning classifier in place of a human-authored fitness function, and we focus on the generation of entire GVGAI games.

\subsection{Autonomous Game Design}

Autonomous game design requires the use of some form of PCG for a minimum of level and rule generation.
Some examples of PCG for autonomous game design involved search-based approaches for board game generation \cite{pell1992metagame, browne2010evolutionary}.
Other early examples used genetic algorithms \cite{cook2012aesthetic}, rule-based conversions from human-authored graphs to games \cite{treanor2012game}, or constraint-solving methods \cite{zook2014automatic}.
More recently some have suggested taking human-authored games and blending the rulesets \cite{gow2015towards}.
However, this concept has not been implemented.
Recently the GVGAI competition added a rule generation track and with it came two example generators, a constructive generator, and a search-based generator \cite{khalifa2017general}.
Neither generator was capable of producing a diverse array of novel games, and for the search-based approach it was specifically noted that the fitness function used did ``not guarantee neither playability nor interestingness''.
Our work attempts to address this by replacing the hand-authored fitness function with a random forest classifier.

\subsection{Procedural Content Generation via Machine Learning}

Procedural Content Generation via Machine Learning (PCGML) can be used for both level generation and rule generation, both of which we attempt in this work \cite{summerville2018procedural}.
Initially PCGML research focused on level generation rather than rule generation or autonomous game design.
In our system, we take inspiration from generative adversarial networks (GANs), which have been used in level generation \cite{hald2020procedural}, for our adversarial training methodology. 
We lack the space to discuss GANs in detail.
Torrado et al. employed a process by which they iteratively trained a GAN on its own output \cite{torrado2020bootstrapping}, we also employ a similar self-supervised approach. 
However, in their case they used a human-authored fitness function to verify generated output, while we attempt to learn a fitness function.

We employ a Random Forest (RF), as a low-data classifier to approximate a ``human-like'' fitness function.
Random Forests have appeared in prior PCGML work \cite{guzdial2018explainable, sarkar2020conditional, sarkar2020sequential}.
The major difference is that in this prior work, the RF is trained once on existing data, while we iteratively train our RF on generated data to approximate our desired fitness function.

While PCGML has largely been applied to level generation, it has also been used for rule generation.
Some researchers have demonstrated the ability to learn game mechanics from human-authored games and apply this to the generation of novel mechanics and levels \cite{Guzdial2017,9356343, summerville2020extracting}. 
There have also been proposals for other complete ML-based automated game design systems \cite{9231927, osborn2017automated}, however, only components of these systems have been implemented at this point \cite{summerville2020extracting}.
Thus far, all of these approaches focus on platformer games, where our system focuses on Video Game Description Language (VGDL) games, which represent a number of game genres.

\section{System Overview}



\begin{figure*}[tbh]
    \centering
    \includegraphics[scale=0.35]{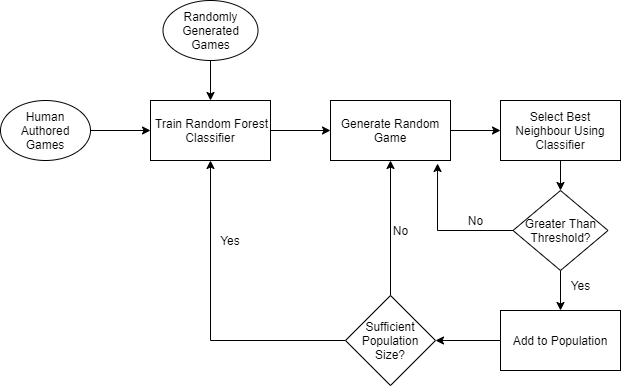}
    \caption{System Flow Chart}
    \label{fig:SystemFlow}
\end{figure*}

In this section we overview our game representation and discuss our novel random forest (RF) classifier-based search method for generating games.
Code for this paper is available on github.\footnote{\url{https://github.com/HonestPretzels/RF_Adversarial_PCG}}
The games that are produced by this system are written in the Video Game Description Language (VGDL) format \cite{schaul2013video}. 
It is the format currently used for the General Video Game Artificial Intelligence competition \cite{perez2019general}, and allows for many kinds of arcade-style, two-dimensional games such as Frogger, Zelda, and Aliens.
We visualize our entire approach in Figure \ref{fig:SystemFlow}, which we'll describe in further detail in subsections below.
It starts by randomly generating a set of VGDL games.
These games are then vectorized into a format usable by our classifier. 
The same vectorization is applied to the set of human-authored VGDL games provided with the library. 
These sets are then labelled and used to train our RF classifier to differentiate between human and random games.
Once the RF is initialized, we use it in place of a fitness function to guide a search-based PCG process to generate new games.
Essentially, the RF's classification acts as an approximated measure of human-likeness for the games in our search space. 
For our experimental implementation, we use a simple greedy hill-climbing search.
We repeatedly generate games using this method until the number of games is equal to the number of human-authored games.
We then use these new round of generated games to retrain our classifier to differentiate between generated and human-authored games. 
Our hope is that this will force the RF to iteratively improve its model for human game classification, which will in turn represent a stronger approximation of a human-like game fitness function.

\subsection{Game Data}

In the VGDL format games are represented by their rules and levels \cite{schaul2013video}.
The rules section contains a description of the sprites in a level, how to handle interactions between those sprites, and the termination conditions which end the game in a win or a loss.
The level is a tile map that can be used to place sprites for the start of the game.
While it is possible to define multiple levels for one set of rules, in our representation we define each distinct game as having a single set of rules and a single level, which increases the small amount of available training data.

\subsection{Random Game Generation}
To create the initial games for our system we randomly generate description and level files.
Each game is created by first selecting the avatar and set of sprites for the game at random.
Some interactions are then randomly selected from the list of possible interactions, and randomly assigned to two sprites from the selected sprite set.
Finally win and loss conditions are chosen at random from the possibilities defined by the VGDL format.
The VGDL library comes with a set of modifiers for rules, such as the speed of sprites or the score change of interactions, and we randomly assign some of these to the generated rules to ensure variation.
Once the rules for a game are generated, we create a level by randomly choosing sprites from our the selected set for each tile in the level, which is set to the most common size of human-authored level.
Before we begin searching each game is tested to ensure that it does not immediately crash by having a random agent play the game for a small number of moves.

Our system also requires a neighbour function to define our search space. 
Our neighbour function generates both rule-based and level-based neighbours. 
To generate a rule-based neighbour we randomly select either a sprite, interaction, or termination rule and replace it by another random rule of the same type.
We did not add or remove rules as we wanted to address rule generation as a separate problem.
To generate a level neighbour we either randomly replace up to ten of the level tiles, or add or delete one of the rows or columns.
These are relatively large changes, but would allow for the system to eventually reach any of the human-authored games as long as the starting random game had the same number of rule types.\par

\subsection{Representation}
To train our random forest (RF) classifier we must change the representation of the games from level and description files into vectors. 
While this representation is novel and unique it follows a standard approach similar to bag-of-words as used in natural language processing.
Our vectorization process starts by counting each sprite, interaction, and termination class present in the description file.
Any classes which are not present in the file are represented as zeros.
We also include information on the sprites involved in each termination condition, as well as the number required for termination.
The representation of the level contains more positional information as levels are inherently positional.
To capture this we convert each character in the level into an integer based on the class of the sprite that it represents and flatten the array into a single dimension vector, padded with zeros for smaller levels.
All of these sections are concatenated into a one dimensional vector of length 413, where the first 53 indices represent the rule information and the remaining represent the level information.

\subsection{Training Approach}
The training approach we use in this system takes some inspiration from generative adversarial networks or GANs.
GANs are a powerful machine learning framework based on having two networks compete with each other.
However, we cannot apply a GAN in our work as there is insufficient training data available.
Instead, we use a RF classifier that attempts to differentiate between human-authored, and generated games, and we retrain it using an adversarial approach.
A RF classifier is a set of decision trees that are initialized stochastically, and fit to a training data set.
When given a test case to evaluate, each decision tree makes a prediction of the class, which is treated as a ``vote'' in determining the final prediction.
It is possible to approximate a probability of a particular class based on the number of trees ``voting'' for that class.
We employ the probability of a game being a human-authored game as our learned fitness function.
Our RF classifier is initially trained with a set of 8 human-authored games, and 8 randomly generated games.
We used the 8 games from the default set of human-authored games as we wanted to replicate real-world cases where acquiring a large backlog of games would be challenging. \par

At every iteration we use our learned fitness function to guide our greedy search.
For any point in the search we select the neighbour with the highest probability of being human-authored, according to our classifier.
Once we have found a game that passes a predefined threshold of 95\% ``human quality'' according to our classifier we save it.
Once we have eight generated games, we retrain the random forest classifier with the new set of generated games and the original eight human-authored games. 
Our goal is that this improves the classifier's ability to discriminate between human-authored and generated games. 

\section{Experiments}
In this section we present two different versions of our generation system, and discuss the results of a series of experiments.

We ran our experiments on two different versions of our system.
The first did not contain any information on the game's termination conditions in the representation, while the second included information on win and loss conditions, as well as the counts that would trigger those conditions.
This was done to determine how the random forests training might differ with and without this information.
Specifically, we hypothesized that without termination conditions it might prove difficult for the random forest (RF) to learn that the human-authored games could all be completed. 
We ran each version of the system for 5 iterations.
While this might seem low in comparison to other ML systems in each iteration we completely retrain the RF, thus it is equivalent to completely retraining a model five times.

Given that the goal of our system was to create games of human-authored quality we had to determine a metric to approximate that quality.
Unfortunately, this can be a rather ambiguous goal and without a human study we decided to evaluate our games based on the relative performance of game playing agents.
This metric has been used in previous research to evaluate game content quality \cite{zook2019montecarlo}.
We modified an implementation of a Monte Carlo tree search (MCTS) agent\footnote{https://github.com/gsurma/slitherin/blob/master/game} designed to play VGDL games and compared its performance against that of a naive random agent.
As MCTS is a more advanced approach, we expect it to outperform the random agent on more complex, human-like games \cite{soemers2016enhancements}.
This expectation is supported by the relative performance of the agents when playing human-authored games, where the MCTS agent was able to complete three more games than the random agent, score one point higher on average (in a scoring range of 0 to 20), and complete the games in an average of 256 fewer moves.

To measure the performance of the agents as they played the games we tracked the following metrics in each iteration of training:
\begin{itemize}
    \item \textbf{Games Completed}: How many generated games an agent completed per iteration.
    \item \textbf{Avg Score}: Average score across the generated games each iteration.
   \item \textbf{Max Score}: Maximum score across the generated games each iteration.
    \item \textbf{Avg Num Moves}: Average number of moves across each game each iteration.
\end{itemize}
\noindent
These metrics allow us to capture the performance of the agents based on criterion that human players might care about, such as scores, high scores, and the number of moves required to win.
During the experiments we limited the agents to five minutes of play time, as well as limited the maximum number of moves to 700.
As the games are simple, they do not require a significant amount of play time. 
These limitations were more than enough for the human-authored games.

\section{Results}

\begin{table*}[tbh]
\small
\begin{tabular}{@{}lllllllll@{}}
\toprule
                 & \multicolumn{2}{l}{Games Completed} & \multicolumn{2}{l}{Avg Score} & \multicolumn{2}{l}{Max Score} & \multicolumn{2}{l}{Avg Num Moves} \\ \midrule
                 & MCTS            & Random            & MCTS          & Random        & MCTS          & Random        & MCTS           & Random           \\
Human   & 6               & 3                 & 5$\pm5$         & 4$\pm7$          & 20            & 15            & 286$\pm263$            & 542$\pm269$              \\
\hline
No-term. 1 & 6               & 6                 & -439$\pm1243$          & -439$\pm1243$          & 4             & 4             & 381$\pm326$            & 385$\pm329$              \\
No-term. 2 & 4               & 4                 & 33315$\pm93252$         & 33249$\pm93118$         & 264085        & 263685        & 374$\pm232$            & 443$\pm292$              \\
No-term. 3 & 4               & 4                 & 5$\pm15$          & 5$\pm15$          & 42            & 42            & 441$\pm277$            & 456$\pm290$              \\
No-term. 4 & 2               & 2                 & -51966$\pm89201$        & -74045$\pm126792$        & 0             & 0             & 379$\pm233$            & 525$\pm234$              \\
No-term. 5 & 2               & 2                 & -5338$\pm15225$         & -3970$\pm10919$         & 755           & 505           & 499$\pm290$            & 410$\pm182$              \\
\hline
Term. 1    & 2               & 2                 & -816$\pm2717$          & -816$\pm2717$          & 960           & 960           & 533$\pm230$            & 595$\pm245$              \\
Term. 2    & 3               & 3                 & 0             & 0             & 0             & 0             & 360$\pm303$            & 438$\pm362$              \\
Term. 3    & 1               & 1                 & 0             & 0             & 0             & 0             & 558$\pm100$            & 666$\pm99$              \\
Term. 4    & 3               & 3                 & 0$\pm1$             & 0$\pm1$             & 3             & 3             & 339$\pm280$            & 438$\pm362$              \\
Term. 5    & 2               & 2                 & 1$\pm11$             & 1$\pm11$             & 26            & 26            & 491$\pm261$            & 550$\pm283$              \\ \bottomrule
\end{tabular}
\caption{Results of experiments for no-termination (no-term.) and with termination (term.) representations.}
\label{tab:results}
\end{table*}

    \begin{figure}[tbh]
        \centering
        \includegraphics[scale=0.08]{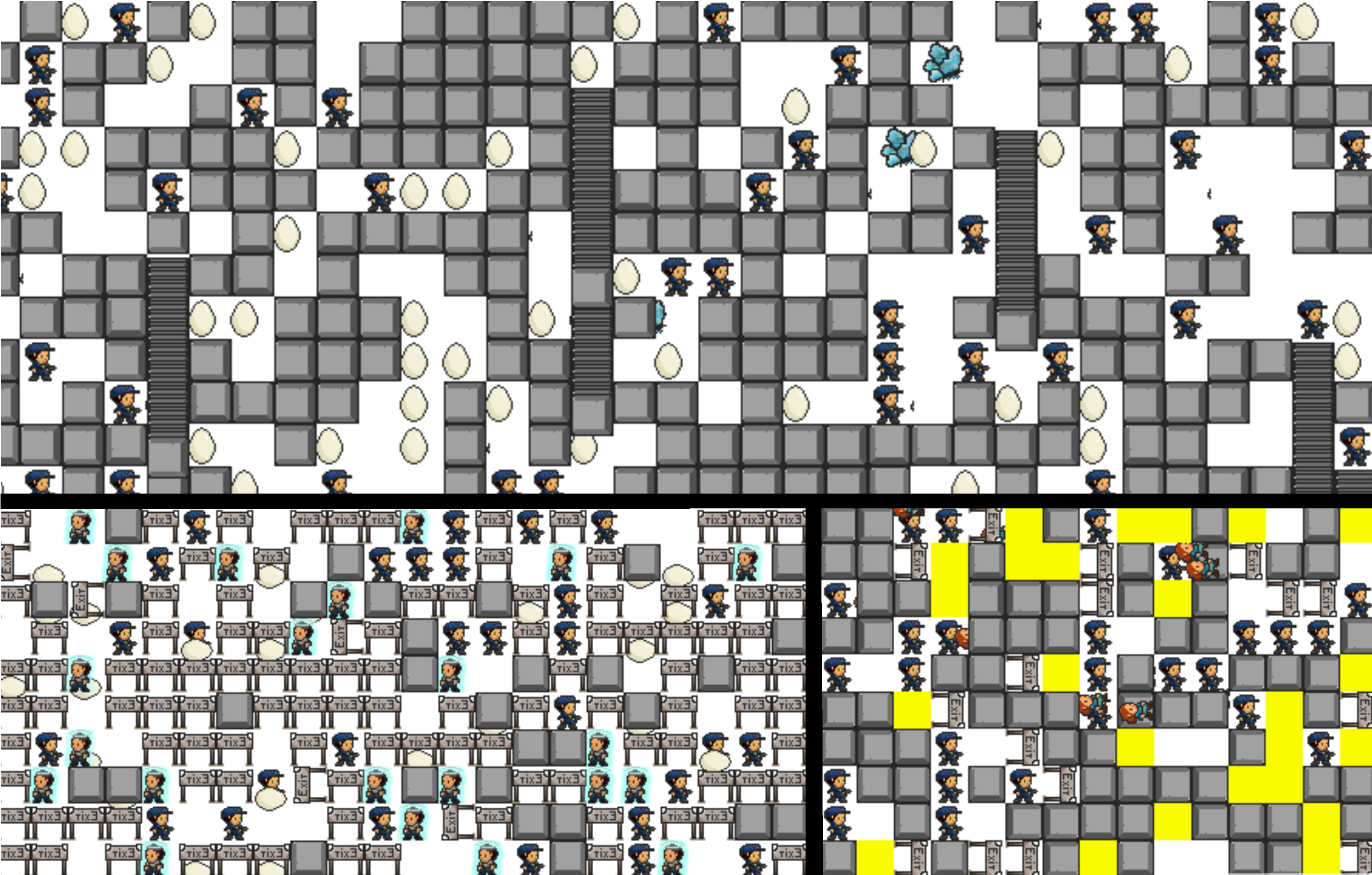}
        \caption{Screenshots of games generated by the No-termination version.}
        \label{fig:versionOneGames}
    \end{figure}
    \begin{figure}[tbh]
        \centering
        \includegraphics[scale=0.08]{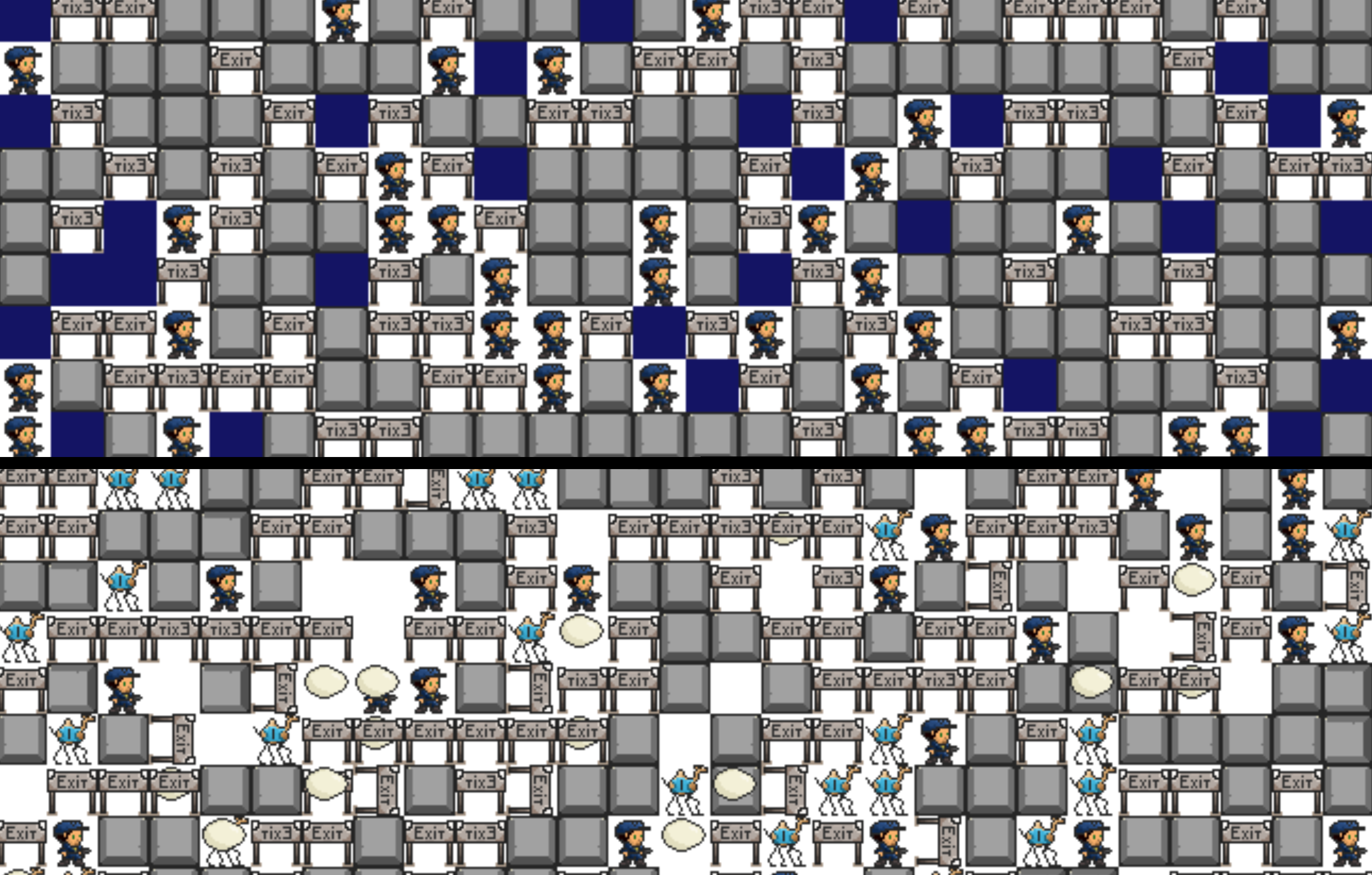}
        \caption{Screenshots of games generated by the Termination version.}
        \label{fig:versionTwoGames}
    \end{figure}

The results of running our experiments are shown in Table \ref{tab:results}.
The first row ``Human-authored'' gives the results of running our two agents on the eight original human-authored games. 
All of the remaining rows give the results of the five iterations for the two versions of our game.

Both agents performed equally well in the games completion metric over every iteration for both versions of our system.
More games were completed in the no-termination version, but most of these games had timer termination conditions and automatically completed after a set amount of time.
We attribute this to the RF learning to classify based on the termination conditions when they were included in our input game vector representation.
This indicates the importance of including this information in order to avoid trivial games.

For the average score metric, only No-termination showed any difference between the agents.
The increase in performance of the MCTS agent on this metric was largest in iteration 4 and dropped off again in the final iteration, a pattern that appears in the other metrics as well.
For example, the difference between MCTS and random in maximum score was greatest in iteration 2 of No-termination and smaller in iteration 5.
Notably, neither version of our system included any information on score change in the game representation, and score changes were randomly assigned to different interactions during generation.
As such the MCTS agent would not always be able to cause the interactions that increase score, or avoid the ones that decrease score.
The RF therefore likely learned to classify on a secondary that indirectly impacted score. 
However, repeatedly retraining appears to have led to the RF ``losing'' this feature, which we'll discuss further below.

In nearly every iteration, across both versions of the system, the MCTS agent played less moves on average than the random agent.
Unfortunately, we believe that this difference was not due to the quality of the games, but rather to the evaluation methodology.
During our experiments we limited the time that each agent had to play the games to five minutes.
Considering the Games Completed metric, the majority of the time the agents stopped playing because of the time limit since, on average, less than half of the eight games generated each iteration were completed.
Due to the relative complexity of MCTS, it played less moves than the random agent in this time.
We believe that this is the primary cause for the apparent increased performance of the MCTS agent versus the random agent.
However, there is still a pattern present in both versions of the system, where the relative performance of the MCTS agent increases towards one of the middle iterations and then begins to drop off.
This is a similar pattern to the one seen for the average and maximum scores.
In effect, in focusing on differentiation of the human games and the later generated games, the classification became less useful as a fitness function.
This follows an intuition that as the generated games became more similar to the human games for some features, the random forest began to pay less attention to those features, as they were no longer helpful for differentiation. 
One answer to this problem might have been to continue to include purely random games in the RF's training data each iteration.

We noticed some interesting qualitative patterns in our generated games.
The games generated with No-termination seemed to have more variety and randomness.
Many of these games varied in level size and sprite set as can be seen in Figure \ref{fig:versionOneGames}.
In comparison, the games generated with Termination had more consistent and noticeable patterns.
For example, in several of the generated games there were octopus sprites which would move and spawn blocks behind them.
This could be because the inclusion of termination information led to fewer possible combinations of rules and levels that could achieve those termination conditions, and thus constrained the search space.
Most of these games had similar sized levels and several of these games had the same sprite groupings, such as many eggs and camels, or many coloured resources, as can be seen in Figure \ref{fig:versionTwoGames}.
It's likely that the classifier learned some relationship between these groupings and ``humaness''.
Almost all the games, across both versions, had a large number of block sprites in their levels.
This occurred as the representation we used did not differentiation between immovable sprites, and the most prevalent immovable sprite in the human-authored games was an immovable background tile.
Additionally, the games included many exit signs, which corresponded to portals/goal blocks in many of the human-authored games.
This indicates that the RF keyed into this sprite as an important indicator of human games.


\section{Future work}

We identified a number of problems above, which in turn suggest directions for future work. 
In terms of the inability to consistently improve over the iterations, we anticipate that this may have been due to a lack of ability to generalize over the available data. 
Increasing the number of human-authored games and/or included random games among the generated games each iteration might address this problem.
From the relative performance of the Termination and No-Termination versions, we identify the importance of representation for this problem.
It would likely be useful to increase the number of sprite-types to better capture designer intent, for example by differentiating between immovable wall sprites, and immovable background sprites.
It could also be useful to increase the information captured on sprite interactions by including which sprites are interacting with each other.
In addition to the improvements suggested for our representation, future work could also focus on the search method itself.
In our work we used a simple greedy search to best interrogate our learned fitness function, but more advanced search methods such as genetic algorithms (GA) could prove useful to better search the space. 
The consistency of games in the Termination condition suggests these were all near a local optima, where a GA might be able to find other optima. 


\section{Conclusions}


We introduced a search-based PCG autonomous game design system using a random forest (RF) classifier in place of a hand-authored fitness function.
We trained our RF in a semi-supervised, adversarial manner.
Our system was unable to produce games of human quality, but the RF clearly learned features of human games.
We provided an analysis of our approach in the hopes of informing future autonomous game design research.

\begin{acks}
This work was funded by the Canada CIFAR AI Chairs Program. We acknowledge the support of the Alberta Machine Intelligence Institute (Amii).
\end{acks}

\bibliographystyle{ACM-Reference-Format}
\bibliography{main}


\begin{thebibliography}{30}


\ifx \showCODEN    \undefined \def \showCODEN     #1{\unskip}     \fi
\ifx \showDOI      \undefined \def \showDOI       #1{#1}\fi
\ifx \showISBNx    \undefined \def \showISBNx     #1{\unskip}     \fi
\ifx \showISBNxiii \undefined \def \showISBNxiii  #1{\unskip}     \fi
\ifx \showISSN     \undefined \def \showISSN      #1{\unskip}     \fi
\ifx \showLCCN     \undefined \def \showLCCN      #1{\unskip}     \fi
\ifx \shownote     \undefined \def \shownote      #1{#1}          \fi
\ifx \showarticletitle \undefined \def \showarticletitle #1{#1}   \fi
\ifx \showURL      \undefined \def \showURL       {\relax}        \fi
\providecommand\bibfield[2]{#2}
\providecommand\bibinfo[2]{#2}
\providecommand\natexlab[1]{#1}
\providecommand\showeprint[2][]{arXiv:#2}

\bibitem[\protect\citeauthoryear{Browne and Maire}{Browne and Maire}{2010}]%
        {browne2010evolutionary}
\bibfield{author}{\bibinfo{person}{Cameron Browne} {and}
  \bibinfo{person}{Frederic Maire}.} \bibinfo{year}{2010}\natexlab{}.
\newblock \showarticletitle{Evolutionary game design}.
\newblock \bibinfo{journal}{\emph{IEEE Transactions on Computational
  Intelligence and AI in Games}} \bibinfo{volume}{2}, \bibinfo{number}{1}
  (\bibinfo{year}{2010}), \bibinfo{pages}{1--16}.
\newblock


\bibitem[\protect\citeauthoryear{Cook, Colton, and Gow}{Cook
  et~al\mbox{.}}{2016}]%
        {cook2016angelina}
\bibfield{author}{\bibinfo{person}{Michael Cook}, \bibinfo{person}{Simon
  Colton}, {and} \bibinfo{person}{Jeremy Gow}.}
  \bibinfo{year}{2016}\natexlab{}.
\newblock \showarticletitle{The angelina videogame design system—part ii}.
\newblock \bibinfo{journal}{\emph{IEEE Transactions on Computational
  Intelligence and AI in Games}} \bibinfo{volume}{9}, \bibinfo{number}{3}
  (\bibinfo{year}{2016}), \bibinfo{pages}{254--266}.
\newblock


\bibitem[\protect\citeauthoryear{Cook, Colton, and Pease}{Cook
  et~al\mbox{.}}{2012}]%
        {cook2012aesthetic}
\bibfield{author}{\bibinfo{person}{Michael Cook}, \bibinfo{person}{Simon
  Colton}, {and} \bibinfo{person}{Alison Pease}.}
  \bibinfo{year}{2012}\natexlab{}.
\newblock \showarticletitle{Aesthetic considerations for automated platformer
  design}. In \bibinfo{booktitle}{\emph{Proceedings of the AAAI Conference on
  Artificial Intelligence and Interactive Digital Entertainment}},
  Vol.~\bibinfo{volume}{8}.
\newblock


\bibitem[\protect\citeauthoryear{Gow and Corneli}{Gow and Corneli}{2015}]%
        {gow2015towards}
\bibfield{author}{\bibinfo{person}{Jeremy Gow} {and} \bibinfo{person}{Joseph
  Corneli}.} \bibinfo{year}{2015}\natexlab{}.
\newblock \showarticletitle{Towards generating novel games using conceptual
  blending}. In \bibinfo{booktitle}{\emph{Proceedings of the AAAI Conference on
  Artificial Intelligence and Interactive Digital Entertainment}},
  Vol.~\bibinfo{volume}{11}.
\newblock


\bibitem[\protect\citeauthoryear{Guzdial, Li, and Riedl}{Guzdial
  et~al\mbox{.}}{2017}]%
        {Guzdial2017}
\bibfield{author}{\bibinfo{person}{Matthew Guzdial}, \bibinfo{person}{Boyang
  Li}, {and} \bibinfo{person}{Mark~O. Riedl}.} \bibinfo{year}{2017}\natexlab{}.
\newblock \showarticletitle{Game Engine Learning from Video}. In
  \bibinfo{booktitle}{\emph{IJCAI}}.
\newblock


\bibitem[\protect\citeauthoryear{Guzdial, Reno, Chen, Smith, and Riedl}{Guzdial
  et~al\mbox{.}}{2018}]%
        {guzdial2018explainable}
\bibfield{author}{\bibinfo{person}{Matthew Guzdial}, \bibinfo{person}{Joshua
  Reno}, \bibinfo{person}{Jonathan Chen}, \bibinfo{person}{Gillian Smith},
  {and} \bibinfo{person}{Mark Riedl}.} \bibinfo{year}{2018}\natexlab{}.
\newblock \showarticletitle{Explainable PCGML via game design patterns}.
\newblock \bibinfo{journal}{\emph{arXiv preprint arXiv:1809.09419}}
  (\bibinfo{year}{2018}).
\newblock


\bibitem[\protect\citeauthoryear{{Guzdial} and {Riedl}}{{Guzdial} and
  {Riedl}}{2021}]%
        {9356343}
\bibfield{author}{\bibinfo{person}{M. {Guzdial}} {and} \bibinfo{person}{M.
  {Riedl}}.} \bibinfo{year}{2021}\natexlab{}.
\newblock \showarticletitle{Conceptual Game Expansion}.
\newblock \bibinfo{journal}{\emph{IEEE Transactions on Games}}
  (\bibinfo{year}{2021}), \bibinfo{pages}{1--1}.
\newblock
\urldef\tempurl%
\url{https://doi.org/10.1109/TG.2021.3060005}
\showDOI{\tempurl}


\bibitem[\protect\citeauthoryear{Hald, Hansen, Kristensen, and Burelli}{Hald
  et~al\mbox{.}}{2020}]%
        {hald2020procedural}
\bibfield{author}{\bibinfo{person}{Andreas Hald},
  \bibinfo{person}{Jens~Struckmann Hansen}, \bibinfo{person}{Jeppe Kristensen},
  {and} \bibinfo{person}{Paolo Burelli}.} \bibinfo{year}{2020}\natexlab{}.
\newblock \showarticletitle{Procedural Content Generation of Puzzle Games using
  Conditional Generative Adversarial Networks}. In
  \bibinfo{booktitle}{\emph{International Conference on the Foundations of
  Digital Games}}. \bibinfo{pages}{1--9}.
\newblock


\bibitem[\protect\citeauthoryear{Khalifa, Green, Perez-Liebana, and
  Togelius}{Khalifa et~al\mbox{.}}{2017}]%
        {khalifa2017general}
\bibfield{author}{\bibinfo{person}{Ahmed Khalifa},
  \bibinfo{person}{Michael~Cerny Green}, \bibinfo{person}{Diego Perez-Liebana},
  {and} \bibinfo{person}{Julian Togelius}.} \bibinfo{year}{2017}\natexlab{}.
\newblock \showarticletitle{General video game rule generation}. In
  \bibinfo{booktitle}{\emph{2017 IEEE Conference on Computational Intelligence
  and Games (CIG)}}. IEEE, \bibinfo{pages}{170--177}.
\newblock


\bibitem[\protect\citeauthoryear{Khalifa, Perez-Liebana, Lucas, and
  Togelius}{Khalifa et~al\mbox{.}}{2016}]%
        {khalifa2016general}
\bibfield{author}{\bibinfo{person}{Ahmed Khalifa}, \bibinfo{person}{Diego
  Perez-Liebana}, \bibinfo{person}{Simon~M Lucas}, {and}
  \bibinfo{person}{Julian Togelius}.} \bibinfo{year}{2016}\natexlab{}.
\newblock \showarticletitle{General video game level generation}. In
  \bibinfo{booktitle}{\emph{Proceedings of the Genetic and Evolutionary
  Computation Conference 2016}}. \bibinfo{pages}{253--259}.
\newblock


\bibitem[\protect\citeauthoryear{{Moghadam} and {Rafsanjani}}{{Moghadam} and
  {Rafsanjani}}{2017}]%
        {Moghadam2017}
\bibfield{author}{\bibinfo{person}{A.~B. {Moghadam}} {and}
  \bibinfo{person}{M.~K. {Rafsanjani}}.} \bibinfo{year}{2017}\natexlab{}.
\newblock \showarticletitle{A genetic approach in procedural content generation
  for platformer games level creation}. In \bibinfo{booktitle}{\emph{2017 2nd
  Conference on Swarm Intelligence and Evolutionary Computation (CSIEC)}}.
  \bibinfo{pages}{141--146}.
\newblock
\urldef\tempurl%
\url{https://doi.org/10.1109/CSIEC.2017.7940160}
\showDOI{\tempurl}


\bibitem[\protect\citeauthoryear{Osborn, Summerville, and Mateas}{Osborn
  et~al\mbox{.}}{2017}]%
        {osborn2017automated}
\bibfield{author}{\bibinfo{person}{Joseph~C Osborn}, \bibinfo{person}{Adam
  Summerville}, {and} \bibinfo{person}{Michael Mateas}.}
  \bibinfo{year}{2017}\natexlab{}.
\newblock \showarticletitle{Automated game design learning}. In
  \bibinfo{booktitle}{\emph{2017 IEEE Conference on Computational Intelligence
  and Games (CIG)}}. IEEE, \bibinfo{pages}{240--247}.
\newblock


\bibitem[\protect\citeauthoryear{Pell}{Pell}{1992}]%
        {pell1992metagame}
\bibfield{author}{\bibinfo{person}{Barney Pell}.}
  \bibinfo{year}{1992}\natexlab{}.
\newblock \showarticletitle{Metagame in symmetric chess-like games}.
\newblock  (\bibinfo{year}{1992}).
\newblock


\bibitem[\protect\citeauthoryear{Perez-Liebana, Liu, Khalifa, Gaina, Togelius,
  and Lucas}{Perez-Liebana et~al\mbox{.}}{2019}]%
        {perez2019general}
\bibfield{author}{\bibinfo{person}{Diego Perez-Liebana},
  \bibinfo{person}{Jialin Liu}, \bibinfo{person}{Ahmed Khalifa},
  \bibinfo{person}{Raluca~D Gaina}, \bibinfo{person}{Julian Togelius}, {and}
  \bibinfo{person}{Simon~M Lucas}.} \bibinfo{year}{2019}\natexlab{}.
\newblock \showarticletitle{General video game ai: A multitrack framework for
  evaluating agents, games, and content generation algorithms}.
\newblock \bibinfo{journal}{\emph{IEEE Transactions on Games}}
  \bibinfo{volume}{11}, \bibinfo{number}{3} (\bibinfo{year}{2019}),
  \bibinfo{pages}{195--214}.
\newblock


\bibitem[\protect\citeauthoryear{Risi and Togelius}{Risi and Togelius}{2020}]%
        {risi2020increasing}
\bibfield{author}{\bibinfo{person}{Sebastian Risi} {and}
  \bibinfo{person}{Julian Togelius}.} \bibinfo{year}{2020}\natexlab{}.
\newblock \showarticletitle{Increasing generality in machine learning through
  procedural content generation}.
\newblock \bibinfo{journal}{\emph{Nature Machine Intelligence}}
  \bibinfo{volume}{2}, \bibinfo{number}{8} (\bibinfo{year}{2020}),
  \bibinfo{pages}{428--436}.
\newblock


\bibitem[\protect\citeauthoryear{Sarkar and Cooper}{Sarkar and Cooper}{2020}]%
        {sarkar2020sequential}
\bibfield{author}{\bibinfo{person}{Anurag Sarkar} {and} \bibinfo{person}{Seth
  Cooper}.} \bibinfo{year}{2020}\natexlab{}.
\newblock \showarticletitle{Sequential segment-based level generation and
  blending using variational autoencoders}. In
  \bibinfo{booktitle}{\emph{International Conference on the Foundations of
  Digital Games}}. \bibinfo{pages}{1--9}.
\newblock


\bibitem[\protect\citeauthoryear{{Sarkar} and {Cooper}}{{Sarkar} and
  {Cooper}}{2020}]%
        {9231927}
\bibfield{author}{\bibinfo{person}{A. {Sarkar}} {and} \bibinfo{person}{S.
  {Cooper}}.} \bibinfo{year}{2020}\natexlab{}.
\newblock \showarticletitle{Towards Game Design via Creative Machine Learning
  (GDCML)}. In \bibinfo{booktitle}{\emph{2020 IEEE Conference on Games (CoG)}}.
  \bibinfo{pages}{744--751}.
\newblock
\urldef\tempurl%
\url{https://doi.org/10.1109/CoG47356.2020.9231927}
\showDOI{\tempurl}


\bibitem[\protect\citeauthoryear{Sarkar, Yang, and Cooper}{Sarkar
  et~al\mbox{.}}{2020}]%
        {sarkar2020conditional}
\bibfield{author}{\bibinfo{person}{Anurag Sarkar}, \bibinfo{person}{Zhihan
  Yang}, {and} \bibinfo{person}{Seth Cooper}.} \bibinfo{year}{2020}\natexlab{}.
\newblock \showarticletitle{Conditional Level Generation and Game Blending}.
\newblock \bibinfo{journal}{\emph{arXiv preprint arXiv:2010.07735}}
  (\bibinfo{year}{2020}).
\newblock


\bibitem[\protect\citeauthoryear{Schaul}{Schaul}{2013}]%
        {schaul2013video}
\bibfield{author}{\bibinfo{person}{Tom Schaul}.}
  \bibinfo{year}{2013}\natexlab{}.
\newblock \showarticletitle{A video game description language for model-based
  or interactive learning}. In \bibinfo{booktitle}{\emph{2013 IEEE Conference
  on Computational Inteligence in Games (CIG)}}. IEEE, \bibinfo{pages}{1--8}.
\newblock


\bibitem[\protect\citeauthoryear{Shaker, Togelius, and Nelson}{Shaker
  et~al\mbox{.}}{2016}]%
        {shaker2016procedural}
\bibfield{author}{\bibinfo{person}{Noor Shaker}, \bibinfo{person}{Julian
  Togelius}, {and} \bibinfo{person}{Mark~J Nelson}.}
  \bibinfo{year}{2016}\natexlab{}.
\newblock \bibinfo{booktitle}{\emph{Procedural content generation in games}}.
\newblock \bibinfo{publisher}{Springer}.
\newblock


\bibitem[\protect\citeauthoryear{Soemers, Sironi, Schuster, and
  Winands}{Soemers et~al\mbox{.}}{2016}]%
        {soemers2016enhancements}
\bibfield{author}{\bibinfo{person}{Dennis~JNJ Soemers},
  \bibinfo{person}{Chiara~F Sironi}, \bibinfo{person}{Torsten Schuster}, {and}
  \bibinfo{person}{Mark~HM Winands}.} \bibinfo{year}{2016}\natexlab{}.
\newblock \showarticletitle{Enhancements for real-time monte-carlo tree search
  in general video game playing}. In \bibinfo{booktitle}{\emph{2016 IEEE
  Conference on Computational Intelligence and Games (CIG)}}. IEEE,
  \bibinfo{pages}{1--8}.
\newblock


\bibitem[\protect\citeauthoryear{Summerville, Martens, Samuel, Osborn,
  Wardrip-Fruin, and Mateas}{Summerville et~al\mbox{.}}{2018a}]%
        {summerville2018gemini}
\bibfield{author}{\bibinfo{person}{Adam Summerville}, \bibinfo{person}{Chris
  Martens}, \bibinfo{person}{Ben Samuel}, \bibinfo{person}{Joseph Osborn},
  \bibinfo{person}{Noah Wardrip-Fruin}, {and} \bibinfo{person}{Michael
  Mateas}.} \bibinfo{year}{2018}\natexlab{a}.
\newblock \showarticletitle{Gemini: Bidirectional generation and analysis of
  games via asp}. In \bibinfo{booktitle}{\emph{Proceedings of the AAAI
  Conference on Artificial Intelligence and Interactive Digital
  Entertainment}}, Vol.~\bibinfo{volume}{14}.
\newblock


\bibitem[\protect\citeauthoryear{Summerville, Sarkar, Snodgrass, and
  Osborn}{Summerville et~al\mbox{.}}{2020}]%
        {summerville2020extracting}
\bibfield{author}{\bibinfo{person}{Adam Summerville}, \bibinfo{person}{Anurag
  Sarkar}, \bibinfo{person}{Sam Snodgrass}, {and} \bibinfo{person}{Joseph
  Osborn}.} \bibinfo{year}{2020}\natexlab{}.
\newblock \showarticletitle{Extracting Physics from Blended Platformer Game
  Levels}. In \bibinfo{booktitle}{\emph{EXAG Workshop}}.
\newblock


\bibitem[\protect\citeauthoryear{Summerville, Snodgrass, Guzdial, Holmgård,
  Hoover, Isaksen, Nealen, and Togelius}{Summerville et~al\mbox{.}}{2018b}]%
        {summerville2018procedural}
\bibfield{author}{\bibinfo{person}{Adam Summerville}, \bibinfo{person}{Sam
  Snodgrass}, \bibinfo{person}{Matthew Guzdial}, \bibinfo{person}{Christoffer
  Holmgård}, \bibinfo{person}{Amy~K. Hoover}, \bibinfo{person}{Aaron Isaksen},
  \bibinfo{person}{Andy Nealen}, {and} \bibinfo{person}{Julian Togelius}.}
  \bibinfo{year}{2018}\natexlab{b}.
\newblock \bibinfo{title}{Procedural Content Generation via Machine Learning
  (PCGML)}.
\newblock
\newblock
\showeprint[arxiv]{1702.00539}~[cs.AI]


\bibitem[\protect\citeauthoryear{{Summerville} and Michael}{{Summerville} and
  Michael}{2016}]%
        {Summerville2016}
\bibfield{author}{\bibinfo{person}{Adam~J. {Summerville}} {and}
  \bibinfo{person}{Mateas Michael}.} \bibinfo{year}{2016}\natexlab{}.
\newblock \showarticletitle{Super Mario as a String: Platformer Level
  Generation Via LSTMs}. In \bibinfo{booktitle}{\emph{DiGRA/FDG \&\#3916 -
  Proceedings of the First International Joint Conference of DiGRA and FDG}}.
  \bibinfo{publisher}{Digital Games Research Association and Society for the
  Advancement of the Science of Digital Games}, \bibinfo{address}{Dundee,
  Scotland}.
\newblock
\showISBNx{ISSN 2342-9666}
\urldef\tempurl%
\url{http://www.digra.org/wp-content/uploads/digital-library/paper_129.pdf}
\showURL{%
\tempurl}


\bibitem[\protect\citeauthoryear{Togelius, Yannakakis, Stanley, and
  Browne}{Togelius et~al\mbox{.}}{2011}]%
        {togelius2011search}
\bibfield{author}{\bibinfo{person}{Julian Togelius},
  \bibinfo{person}{Georgios~N Yannakakis}, \bibinfo{person}{Kenneth~O Stanley},
  {and} \bibinfo{person}{Cameron Browne}.} \bibinfo{year}{2011}\natexlab{}.
\newblock \showarticletitle{Search-based procedural content generation: A
  taxonomy and survey}.
\newblock \bibinfo{journal}{\emph{IEEE Transactions on Computational
  Intelligence and AI in Games}} \bibinfo{volume}{3}, \bibinfo{number}{3}
  (\bibinfo{year}{2011}), \bibinfo{pages}{172--186}.
\newblock


\bibitem[\protect\citeauthoryear{Torrado, Khalifa, Green, Justesen, Risi, and
  Togelius}{Torrado et~al\mbox{.}}{2020}]%
        {torrado2020bootstrapping}
\bibfield{author}{\bibinfo{person}{Ruben~Rodriguez Torrado},
  \bibinfo{person}{Ahmed Khalifa}, \bibinfo{person}{Michael~Cerny Green},
  \bibinfo{person}{Niels Justesen}, \bibinfo{person}{Sebastian Risi}, {and}
  \bibinfo{person}{Julian Togelius}.} \bibinfo{year}{2020}\natexlab{}.
\newblock \showarticletitle{Bootstrapping conditional gans for video game level
  generation}. In \bibinfo{booktitle}{\emph{2020 IEEE Conference on Games
  (CoG)}}. IEEE, \bibinfo{pages}{41--48}.
\newblock


\bibitem[\protect\citeauthoryear{Treanor, Blackford, Mateas, and
  Bogost}{Treanor et~al\mbox{.}}{2012}]%
        {treanor2012game}
\bibfield{author}{\bibinfo{person}{Mike Treanor}, \bibinfo{person}{Bryan
  Blackford}, \bibinfo{person}{Michael Mateas}, {and} \bibinfo{person}{Ian
  Bogost}.} \bibinfo{year}{2012}\natexlab{}.
\newblock \showarticletitle{Game-o-matic: Generating videogames that represent
  ideas}. In \bibinfo{booktitle}{\emph{Proceedings of the The third workshop on
  Procedural Content Generation in Games}}. \bibinfo{pages}{1--8}.
\newblock


\bibitem[\protect\citeauthoryear{Zook, Harrison, and Riedl}{Zook
  et~al\mbox{.}}{2019}]%
        {zook2019montecarlo}
\bibfield{author}{\bibinfo{person}{Alexander Zook}, \bibinfo{person}{Brent
  Harrison}, {and} \bibinfo{person}{Mark~O. Riedl}.}
  \bibinfo{year}{2019}\natexlab{}.
\newblock \bibinfo{title}{Monte-Carlo Tree Search for Simulation-based Strategy
  Analysis}.
\newblock
\newblock
\showeprint[arxiv]{1908.01423}~[cs.AI]


\bibitem[\protect\citeauthoryear{Zook and Riedl}{Zook and Riedl}{2014}]%
        {zook2014automatic}
\bibfield{author}{\bibinfo{person}{Alexander Zook} {and} \bibinfo{person}{Mark
  Riedl}.} \bibinfo{year}{2014}\natexlab{}.
\newblock \showarticletitle{Automatic game design via mechanic generation}. In
  \bibinfo{booktitle}{\emph{Proceedings of the AAAI Conference on Artificial
  Intelligence}}, Vol.~\bibinfo{volume}{28}.
\newblock


\end{thebibliography}

\appendix

\end{document}